\pdfoutput=1

\documentclass[11pt]{article}

\usepackage[]{ACL2023}

\usepackage{times}
\usepackage{latexsym}

\usepackage[T1]{fontenc}

\usepackage[utf8]{inputenc}

\usepackage{microtype}

\usepackage{inconsolata}

\usepackage{amsfonts}
\usepackage{amsmath}
\usepackage{bm}
\usepackage{tabularx}
\usepackage{url}
\usepackage{multirow}
\usepackage{booktabs}
\usepackage{graphicx}
\usepackage{booktabs}
\usepackage{floatrow}
\usepackage{hyperref}

\title{How does the task complexity of masked pretraining objectives affect downstream performance?}

\author{
  Atsuki Yamaguchi{\rm \textsuperscript{1}}{\rm ,} Hiroaki Ozaki{\rm \textsuperscript{1}}$^{\text{*}}${\rm ,} Terufumi Morishita{\rm \textsuperscript{1}}$^{\text{*}}${\rm ,}\\
  {\bf Gaku Morio}{\rm \textsuperscript{2}}$^{\text{*}}$~ and {\bf Yasuhiro Sogawa}{\rm \textsuperscript{1}}\\
  \textsuperscript{1}{Hitachi,  Ltd., Kokubunji, Tokyo, Japan}\\
  \textsuperscript{2}{Hitachi America Ltd., Santa Clara, CA, USA}\\
  \textsuperscript{1}\texttt{\{atsuki.yamaguchi.xn,hiroaki.ozaki.yu,}\\\texttt{terufumi.morishita.wp,yasuhiro.sogawa.tp\}@hitachi.com}\\
  \textsuperscript{2}\texttt{gaku.morio@hal.hitachi.com}\\
}

\begin{document}
\maketitle

{
\let\thefootnote\relax\footnotetext{$^{\text{*}}$ Equal contribution}
}

\begin{abstract}
Masked language modeling (MLM) is a widely used self-supervised pretraining objective, where a model needs to predict an original token that is replaced with a mask given contexts. Although simpler and computationally efficient pretraining objectives, e.g., predicting the first character of a masked token, have recently shown comparable results to MLM, no objectives with a masking scheme actually outperform it in downstream tasks.
Motivated by the assumption that their lack of complexity plays a vital role in the degradation, we validate whether more complex masked objectives can achieve better results and investigate how much complexity they should have to perform comparably to MLM.
Our results using GLUE, SQuAD, and Universal Dependencies benchmarks demonstrate that more complicated objectives tend to show better downstream results with at least half of the MLM complexity needed to perform comparably to MLM. Finally, we discuss how we should pretrain a model using a masked objective from the task complexity perspective.\footnote{Our code and pretrained models are available at \url{https://github.com/hitachi-nlp/mlm-probe-acl2023}.}
\end{abstract}

\section{Introduction} \label{sec:introduction}
Masked language modeling (MLM) \cite{devlin-etal-2019-bert}, where a model needs to predict a particular token that is replaced with a mask placeholder given its surrounding context, is a widely used self-supervised pretraining objective in natural language processing.
Recently, simpler pretraining objectives have shown promising results on downstream tasks. \citet{aroca-ouellette-rudzicz-2020-losses} have proposed various token-level and sentence-level auxiliary pretraining objectives, showing improvements over BERT \cite{devlin-etal-2019-bert}. \citet{yamaguchi-etal-2021-frustratingly} and \citet{alajrami-aletras-2022-pre} have demonstrated that such token-level objectives themselves, i.e., pretraining without MLM, perform comparably to MLM.

Although these simple token-level objectives themselves, e.g., predicting the first character of a masked token (First Char) \cite{yamaguchi-etal-2021-frustratingly}, have exhibited competitive downstream performances to MLM with smaller computations, no objectives using mask tokens are not clearly comparable to MLM on downstream tasks. We conjecture that the main reason behind the performance difference lies in its lack of complexity, i.e., the number of classes to be predicted, and similar arguments have been made for auxiliary task ineffectiveness \cite{Lan2020ALBERT} and pretraining task design \cite{yamaguchi-etal-2021-frustratingly}.

This paper sheds light on the task complexity of masked pretraining objectives and investigates \textbf{RQ1}: whether a more complex objective, becoming closer to MLM, can achieve a better downstream result and \textbf{RQ2}: how much complexity they need to obtain comparable results to MLM.
To this end, we propose masked $n$ character prediction as a control task, which requires us to predict the first or last $n$ characters of a masked token, allowing us to empirically evaluate how the task complexity affects downstream performance by varying $n$.
We pretrain 14 different types of models with the proposed control task in addition to MLM for reference and evaluate their downstream performance on the GLUE \cite{wang2018glue}, SQuAD \cite{rajpurkar-etal-2016-squad}, and Universal Dependencies (UD) \cite{nivre-etal-2020-universal} benchmarks.
We also conduct a cost-benefit analysis of performance gains with respect to task complexity and analyze how to select an optimal complexity for a given task. 

\paragraph{Contributions} 
(1) We model the task complexity of a masked pretraining objective as masked $n$ character prediction (\S\ref{sec:task}) and revealed how it affects downstream performance (\S\ref{sec:results}).
(2) We conduct two analyses to provide insights into how we should pretrain a model by using a masked objective from the task complexity perspective (\S\ref{sec:analysis}).

\section{Methodology} \label{sec:task}
Our main hypothesis is that the more complexity, i.e., the number of classes to be predicted, masked pretraining objectives have, the better the downstream performance they will achieve.
This is because a more complex task should have a larger number of classes to be predicted, giving more informative signals to a model via training.

To verify the hypothesis empirically and answer the two research questions listed in \S\ref{sec:introduction}, we extend First Char \cite{yamaguchi-etal-2021-frustratingly} and let a model predict the first or last $n \in \mathbb{N}$ characters of a masked token ($n$ Chars)\footnote{We also allow \textit{last} $n$ characters to be predicted because the prediction direction should not matter for a model to learn effective representations.}.
The task is trained with the token-level cross-entropy loss averaged over masked tokens. Our extension allows us to evaluate how the complexity affects downstream performance by varying $n$.

\begin{figure}[t!]
    \begin{center}
    \includegraphics[width=\linewidth]{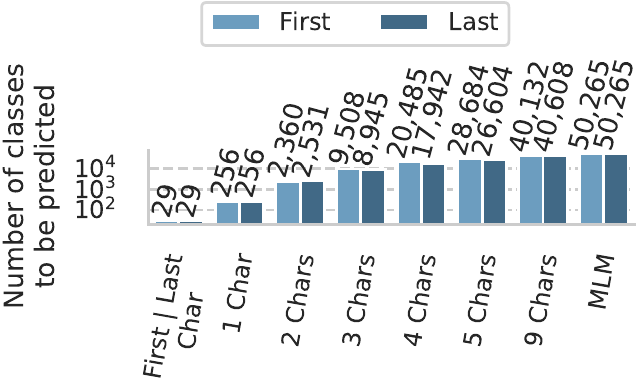}
    \caption{Number of classes to be predicted for each $n$ Chars objective along with First Char, Last Char, and MLM.}
    \label{fig:num_classes}
    \end{center}
\end{figure}

Figure \ref{fig:num_classes} shows the number of classes to be predicted for $n$ Chars when using a pretrained tokenizer of RoBERTa \cite{roberta}. We can see that the larger $n$, the closer the objective becomes to MLM.
For 1 Char, we simply pick up the first character of each token in the vocabulary instead of casting it into 29 classes as in First Char\footnote{In First Char, a model needs to predict the first character of a masked token as 29-way classification, including alphanumeric characters, punctuation marks, and any other characters.}, resulting in 256 types of characters.

\section{Experimental Setup}
Here, we describe our experimental setups for both pretraining and fine-tuning.\footnote{For more details, please refer to Appendix \ref{sec:hyperparameters}.}

\paragraph{Model}
We used the base configuration of BERT \cite{devlin-etal-2019-bert}.
The model consists of 12 hidden layers and attention heads, and the dimensions of hidden layers and intermediate feed-forward layers are 768 and 3072, respectively.
We simply put a linear layer on the BERT model for $n$ Chars and First Char.

\paragraph{Baselines}
We pretrained MLM and First Char for reference to evaluate the influence of masked pretraining objective complexity.
We also set up Last Char, where a model needs to predict the last character of a masked token from 29 categories the same as in First Char.

\renewcommand*{\arraystretch}{1.0}
\begin{table*}[!t]
\begin{center}
\resizebox{\linewidth}{!}{%
\begin{tabular}{lcccccccc|c|c|c}
\toprule
 & \textbf{MNLI} & \textbf{QQP} & \textbf{QNLI} & \textbf{SST} & \textbf{CoLA} & \textbf{STS} & \textbf{MRPC} & \textbf{RTE} & \textbf{GLUE Avg.} & \multicolumn{1}{c|}{\textbf{SQuAD}} & \multicolumn{1}{c}{\textbf{UD}}\\ 
 \textbf{Model} & 393k & 364k & 105k & 67k & 8.6k & 5.7k & 3.7k & 2.5k &  & \multicolumn{1}{c|}{88k} & \multicolumn{1}{c}{13k} \\ 
\midrule
\enskip MLM & \textbf{82.3} & \textbf{86.9} & 89.2 & \underline{91.8} & \textbf{58.0} & \textbf{87.0} & \textbf{86.7} & \textbf{64.8} & \textbf{80.8} (0.3) & \underline{88.1} (0.6) & \textbf{88.8} (0.1)\\
\midrule
\enskip First 9 Chars & 81.6 & 86.4 & 89.2 & \textbf{91.9} & 53.0 & 85.6 & 85.2 & 58.2 & 78.9 (0.9) & 87.4 (0.4) & 88.5 (0.1)\\
\enskip First 5 Chars & 82.0 & 86.6 & 89.3 & 91.1 & 51.8 & 85.6 & 85.5 & 59.2 & 78.9 (0.5) & 87.9 (0.5) & 88.5 (0.1)\\
\enskip First 4 Chars & 82.0 & 86.6 & \textbf{89.6} & 91.3 & 54.2 & 85.5 & 85.7 & 57.3 & 79.0 (0.4) & 87.9 (0.4) & \textbf{88.8} (0.1)\\
\enskip First 3 Chars & 81.9 & \underline{86.8} & 88.7 & 90.7 & 52.0 & 85.9 & 85.6 & 58.9 & 78.8 (0.5) & 87.6 (0.3) & 88.1 (0.1)\\
\enskip First 2 Chars & 81.1 & 86.5 & 88.6 & 90.8 & 51.1 & 85.1 & 83.7 & 60.6 & 78.4 (0.7) & 86.8 (0.7) & 87.8 (0.1)\\
\enskip First 1 Char & 80.5 & 86.3 & 88.5 & 90.4 & 48.6 & 84.7 & 83.3 & 60.0 & 77.8 (0.3) & 86.1 (0.3) & 87.6 (0.1)\\
\enskip First Char & 80.7 & 86.3 & 88.2 & 90.6 & 50.0 & 85.1 & 85.4 & 59.5 & 78.2 (0.3) & 85.6 (0.4) & 87.8 (0.0)\\
\midrule
\enskip Last 9 Chars & \underline{82.1} & 86.7 & 89.3 & 91.4 & 55.0 & 85.6 & 85.1 & 57.6 & 79.1 (0.3) & \textbf{88.4} (0.2) & \underline{88.7} (0.1)\\
\enskip Last 5 Chars & 81.8 & 86.4 & 89.1 & 91.3 & 54.8 & \underline{85.8} & 85.4 & 58.7 & \underline{79.2} (0.8) & 87.5 (0.7) & 88.5 (0.1)\\
\enskip Last 4 Chars & 81.6 & 86.6 & \underline{89.4} & 90.2 & \underline{56.0} & 85.6 & \underline{86.2} & 56.9 & 79.1 (0.3) & 87.6 (0.8) & 88.4 (0.1)\\
\enskip Last 3 Chars & 81.3 & 86.4 & 88.9 & 91.0 & 53.0 & 84.9 & 84.9 & 56.1 & 78.3 (0.6) & 87.0 (0.5) & 88.4 (0.1)\\
\enskip Last 2 Chars & 81.0 & 86.3 & 88.0 & 90.7 & 50.7 & 84.5 & 85.7 & 58.6 & 78.2 (0.4) & 87.0 (0.3) & 88.0 (0.1)\\
\enskip Last 1 Char & 80.2 & 86.3 & 87.9 & 90.4 & 54.8 & 84.6 & 84.8 & \underline{61.2} & 78.8 (0.3) & 86.0 (0.8) & 87.8 (0.1)\\
\enskip Last Char & 79.8 & 86.0 & 87.5 & 90.2 & 48.8 & 85.2 & 85.2 & 55.7 & 77.3 (0.5) & 85.5 (0.1) & 88.1 (0.1)\\
\bottomrule\\
\toprule
\textbf{Correlation} $r$ & \textbf{MNLI} & \textbf{QQP} & \textbf{QNLI} & \textbf{SST} & \textbf{CoLA} & \textbf{STS} & \textbf{MRPC} & \textbf{RTE} & \textbf{GLUE Avg.} & \multicolumn{1}{c|}{\textbf{SQuAD}} & \multicolumn{1}{c}{\textbf{UD}}\\ 
\midrule
First Char/MLM & .787 & .520 & .659 & .734 & .705 & .653 & .486 & .182 & \multicolumn{1}{c|}{.721} & \multicolumn{1}{c|}{.783} & \multicolumn{1}{c}{.873} \\
Last Char/MLM & .942 & .739 & .835 & .662 & .675 & .714 & .322 & .336 & \multicolumn{1}{c|}{.789} & \multicolumn{1}{c|}{.840} & \multicolumn{1}{c}{.877} \\
\bottomrule

\end{tabular}%
}
\caption{Results and their correlation values on GLUE, SQuAD, and UD (EN-EWT) with standard deviations over five runs in parentheses.
Values under dataset names are the number of their corresponding training samples.
We show test set results for UD and dev sets results for GLUE and SQuAD. \textbf{Bold} and \underline{underlined} values denote best and second best scores for each dataset.}
\label{table:base_result}
\end{center}
\end{table*}

\paragraph{Pretraining Data}
Following \citet{devlin-etal-2019-bert}, we pretrained all models on English Wikipedia and BookCorpus \cite{Zhu_2015_ICCV} using the datasets \cite{lhoest-etal-2021-datasets} library.
We set the maximum sequence length to 512.
We tokenized texts using byte-level Byte-Pair-Encoding \cite{sennrich-etal-2016-neural}, and the resulting corpora consist of 10 million samples and 4.9 billion tokens in total.

\paragraph{Fine-tuning Data}
We used the GLUE, SQuAD v1.1, and UD v2.10 benchmarks to measure both semantic and syntactic downstream performances in detail.
For UD, we used its English subset of Universal Dependencies English Web Treebank (EN-EWT) \cite{silveira-etal-2014-gold}.  

\paragraph{Evaluation}
Following previous work \cite{aroca-ouellette-rudzicz-2020-losses}, we report matched accuracy for MNLI, Matthews correlation for CoLA, Spearman correlation for STS-B, accuracy for MRPC, F1 scores for QQP and SQuAD, and accuracy for all other tasks.
For UD, we used labeled attachment score (LAS). For each task, we report a mean score over five runs with different random seeds.
We excluded problematic WNLI following prior work \cite{aroca-ouellette-rudzicz-2020-losses}.

\paragraph{Implementation Details}
We implemented our models using the PyTorch \cite{NEURIPS2019_9015} and Hugging Face Transformers \cite{wolf-etal-2020-transformers} libraries.
For fine-tuning on UD, we trained a deep biaffine attention parser (BAP) \cite{dozat2017deep} built on top of pretrained language models.
We used the SuPar library\footnote{\url{https://github.com/yzhangcs/parser}} to implement the parser and followed its default hyperparameter configurations.

\section{Results} \label{sec:results}
\paragraph{RQ1: Do more complex objectives achieve better results?}
Table \ref{table:base_result} displays downstream task results on GLUE, SQuAD, and UD for our control tasks and their comparisons against MLM, First Char, and Last Char.
Overall, we observe the larger $n$, the better downstream performance is.
Looking at each dataset result closely, we see that the datasets with over 5k training samples exhibit moderate to high correlations with the lowest and highest correlation values of 0.520 (First Char to MLM) for QQP and 0.942 (Last Char to MLM) for MNLI, respectively. In contrast, the corpora with less than 5k samples tend to exhibit low to moderate correlations, ranging from 0.182 (First Char to MLM) for RTE to 0.486 (First Char to MLM) for MRPC.
Therefore, we can see a general trend that a more complex masked pretraining objective yields better performance in downstream tasks especially under a high-resource scenario, whereas it does not always achieve a better result on a low-resource dataset, where in this case MLM tends to achieve a better result.

\paragraph{RQ2: How much complexity do we need to obtain comparable results to MLM?}
To answer the RQ2, we only compare results on high-resource corpora, given that results on low-resource corpora have large standard deviations of 1.0 or more (see Appendix \ref{sec:glue}).
We can observe from Table \ref{table:base_result} that at least $n = 4$ complexity is necessary to achieve comparable results to MLM. For instance, First $n$ Chars objectives require $n = 4$ (20k classes) to surpass the MLM performance on at least one of the target downstream tasks.
Last $n$ Chars also need $n = 4$ (18k classes) to beat MLM on one of the tasks.
For MNLI and QQP, our control tasks did not yield better results.

\renewcommand*{\arraystretch}{1.0}
\begin{table}[!t]
\begin{center}
\tabcolsep 5pt
\resizebox{\linewidth}{!}{%
\begin{tabular}{llllll}
\toprule
 & \textbf{FLOPs} & \textbf{GLUE} & \textbf{SQuAD} & \textbf{UD} \\ 
\textbf{Model} & {$\times 10^{19}$} & {Avg.} & {F1} & {LAS} \\
\midrule
MLM & 2.44 & 80.8 & 88.1 & 88.8 \\
\midrule
\textbf{First}\\
9 Chars & 2.29 (-6) & 78.9 (-2.4) & 87.4 (-0.9) & 88.5 (-0.3)\\
5 Chars & 2.12 (-13) & 78.9 (-2.4) & 87.9 (-0.3) & 88.5 (-0.4)\\
4 Chars & 1.99 (-18) & 79.0 (-2.3) & 87.9 (-0.2) & 88.8 (0.0)\\
3 Chars & 1.83 (-25) & 78.8 (-2.5) & 87.6 (-0.6) & 88.1 (-0.8)\\
2 Chars & 1.72 (-30) & 78.4 (-3.0) & 86.8 (-1.5) & 87.8 (-1.1)\\
1 Char & 1.69 (-31) & 77.8 (-3.8) & 86.1 (-2.3) & 87.6 (-1.4)\\
First Char & 1.68 (-31) & 78.2 (-3.2) & 85.6 (-2.9) & 87.8 (-0.2)\\
\midrule
\textbf{Last} \\
9 Chars & 2.30 (-6) & 79.1 (-2.2) & 88.4 (+0.3) & 88.7 (-0.2)\\
5 Chars & 2.09 (-15) & 79.2 (-2.1) & 87.5 (-0.7) & 88.5 (-0.4)\\
4 Chars & 1.96 (-20) & 79.1 (-2.2) & 87.6 (-0.6) & 88.4 (-0.4)\\
3 Chars & 1.82 (-26) & 78.3 (-3.1) & 87.0 (-1.3) & 88.4 (-0.5)\\
2 Chars & 1.72 (-30) & 78.2 (-3.3) & 87.0 (-1.3) & 88.0 (-1.0)\\
1 Char & 1.69 (-31) & 78.8 (-2.5) & 86.0 (-2.4) & 87.8 (-1.1)\\
Last Char & 1.68 (-31) & 77.3 (-4.4) & 85.5 (-3.0) & 88.1 (-0.8)\\
\bottomrule
\end{tabular}%
}
\caption{Computational efficiency comparison. Values in parentheses are in percent and show relative performance differences from MLM results.}
\label{table:computational_resource_result}
\end{center}
\end{table}

\section{Analysis and Discussion} \label{sec:analysis}
On the basis of results from the RQ1 and RQ2, we discuss how we should pretrain a model in practice from the task complexity perspective.

\paragraph{Task complexity affects computational efficiency.}
The task complexity is closely related to computational costs with more complex objectives larger costs. Table \ref{table:computational_resource_result} shows the number of floating-point operations (FLOPs)\footnote{We use FLOPs to evaluate computational costs following \citet{Clark2020ELECTRA:}.} required for $n$ Chars, First Char, Last Char, and MLM along with its downstream performance.
We can see a clear trade-off between computational efficiency and downstream performance.
Smaller $n$ drastically reduces FLOPs with the maximum relative reduction of 31\% for First Char and Last Char, while larger $n$ has small reduction rates with the minimum value of 6\% for First and Last 9 Chars.
To obtain comparable results to MLM, as we discussed in the RQ2, we need at least $n=4$ (First and Last) complexity, which can only reduce 18\% and 20\% of FLOPs, respectively.
These results suggest that we need careful cost-benefit consideration when using a masked pretraining objective on the basis of target performance and computational costs.

\begin{table}[!t]
\begin{center}
\tabcolsep 4pt
\small
\begin{tabular}{llll}
\toprule
 & \multicolumn{2}{c}{\textbf{F1}$\uparrow$} & \textbf{Miss}$\downarrow$ \\
\textbf{Model} & w/ Miss & w/o Miss & \\
\midrule
MLM & 88.1 & 94.4 & 6.6\\
\midrule
\textbf{First}\\
9 Chars & 87.4 (-0.9) & 94.2 (-0.2) & 7.2 (+9.8)\\
5 Chars & 87.9 (-0.3) & 94.0 (-0.3) & 6.5 (-0.8)\\
4 Chars & 87.9 (-0.2) & 94.3 (0.0) & 6.8 (+3.1)\\
3 Chars & 87.6 (-0.6) & 94.3 (0.0) & 7.1 (+8.1)\\
2 Chars & 86.8 (-1.5) & 93.8 (-0.6) & 7.4 (+12.4)\\
1 Char & 86.1 (-2.3) & 93.9 (-0.5) & 8.2 (+25.0)\\
First Char & 85.6 (-2.9) & 93.9 (-0.5) & 8.8 (+34.0)\\
\midrule
\textbf{Last} \\
9 Chars & 88.4 (+0.3) & 94.3 (-0.1) & 6.3 (-5.1)\\
5 Chars & 87.5 (-0.7) & 94.2 (-0.2) & 7.0 (+6.7)\\
4 Chars & 87.6 (-0.6) & 94.1 (-0.3) & 6.8 (+3.7)\\
3 Chars & 87.0 (-1.3) & 94.0 (-0.4) & 7.5 (+13.1)\\
2 Chars & 87.0 (-1.3) & 94.0 (-0.4) & 7.5 (+13.7)\\
1 Char & 86.0 (-2.4) & 93.7 (-0.7) & 8.2 (+24.2)\\
Last Char & 85.5 (-3.0) & 94.1 (-0.3) & 9.1 (+38.2)\\
\bottomrule
\end{tabular}%
\caption{Results on the SQuAD dev set. ``Miss'' denotes the percentage of samples with no overlap between predicted and gold spans. Values in parentheses are in percent and show relative performance differences from MLM results.}
\label{table:case_study}
\end{center}
\end{table}

\paragraph{How can we select an optimal complexity?}
Finally, we discuss how we can decide an optimal complexity for a specific downstream task by analyzing how the task complexity affects the performance in detail.
Here, we take SQuAD as an example case, where we observed a large relative difference of 3.0\% in Table \ref{table:computational_resource_result}.
Table \ref{table:case_study} lists the results on SQuAD, including the ratio of mis-detection, i.e., the percentage of samples with no overlap between predicted and gold spans, and F1 scores calculated without mis-detected cases.
We found that simple masked objectives were likely to suffer from mis-detection with the worst performance degradation of 38.2\% for Last Char, which is far larger than those observed in other metrics and corpora (see Tables \ref{table:computational_resource_result} and \ref{table:base_result_performance_degradation} in Appendix). In contrast, the relative performance difference values of F1 scores computed without mis-detected samples show quite similar trends to other high-resource corpora, which are typically less than 2\% at a maximum.
These results imply that the task complexity mainly contributes to an increase/decrease in the number of mis-detections in SQuAD, and selecting a complex masked objective (e.g., First 5 Chars) is a safeguard option to minimize the effect of mis-ditection.
Therefore, different downstream tasks might have different optimal complexities due to their characteristics and evaluation metrics, as observed in the example case above. We leave thorough investigation of these effects as future work.

\section{Conclusion}
This paper analyzed the impact of masked objective complexities over downstream performance, motivated by the assumption that the lack of task complexity in simple masked pretraining objectives (e.g., First Char) affects the performance degradation compared to MLM.
Experiments using the GLUE, SQuAD, and UD datasets revealed that the task complexity significantly affected downstream performance with at least 35.7\% of the MLM prediction classes needed to perform comparably to MLM on at least one of the high-resource corpora. Our analysis also showed that there exists a trade-off between downstream performance and computational efficiency, and different downstream tasks might have different optimal complexities.
Future work includes analyzing other properties (e.g., fairness) with respect to task complexity.

\section*{Limitations}
\paragraph{Model Architecture}
Due to our computational resource constraints, we only used the BERT base architecture. We cannot confirm whether our results and observations are transferable to any other Transformer-based architectures, especially for larger ones.

\paragraph{Randomness}
We did not run pretraining for multiple times with different random seeds due to the limited computational resources and research budgets, though we fine-tuned models five times each with different random seeds in any downstream tasks. This might affect the overall results shown in the paper.

\paragraph{Languages Other than English}
It is uncertain whether any results and conclusions presented in this paper are applicable to any other languages other than English, as our experiments are solely on English data. We may need further experiments especially for languages that do not belong to the same language family as English, such as Chinese and Japanese.

\section*{Ethics Statement}
This work does not involve any sensitive data but only uses publicly available data, including Wikipedia, GLUE, SQuAD, and UD as explained in the paper. Although we plan to release the resulting models, they might perform unfairly in some circumstances, as reported in \citet{baldini-etal-2022-fairness}. We highly recommend users to refer to studies on debiasing pretrained language models, such as \citet{guo-etal-2022-auto}.

\section*{Acknowledgements}
We would like to thank anonymous ACL 2023 reviewers and Yuichi Sasazawa for their insightful comments.
We also would like to thank Dr. Masaaki Shimizu for the maintenance and management of the large computational resources used in this paper.

\bibliography{anthology,custom}
\bibliographystyle{acl_natbib}

\newpage
\appendix
\begin{table}[h]
\begin{center}
\small
\begin{tabular}{lc}
\toprule
\textbf{Hyperparameters} & \textbf{Values} \\
\midrule
Batch size & 128 \\
Total training steps & 500,000 \\
Adam $\epsilon$ & 1e-6 \\
Adam $\beta_1$ & 0.9 \\
Adam $\beta_2$ & 0.999 \\
Sequence length & 512 \\
Learning rate & 1e-4 for Last Char \\
& 2e-4 for other models \\
Learning rate schedule & linear warmup \\
Warmup steps & 10,000 \\
Weight decay & 0.01 \\
Attention dropout & 0.1 \\
Dropout & 0.1 \\
\bottomrule
\end{tabular}%
\caption{Hyperparameters for pretraining.}
\label{table:hyperparams_pretraining}
\end{center}
\end{table}

\begin{table*}[t]
\begin{center}
\begin{tabular}{lccc}
\toprule
\textbf{Hyperparameters} & \textbf{GLUE} & \textbf{SQuAD} & \textbf{UD} \\
\midrule
Batch size & 32 & 24 & 32\\
Maximum number of epochs & 20 & 10 & 10\\
Adam $\epsilon$ & 1e-8 & 1e-8 & 1e-8\\
Adam $\beta_1$ & 0.9 & 0.9 & 0.9\\
Adam $\beta_2$ & 0.999 & 0.999 & 0.999\\
Sequence length & 128 & 384 & 512\\
Learning rate & 3e-5 & 3e-5 & 5e-5 for BERT, 1e-3 for BAP \\
Learning rate schedule & linear warmup & linear warmup & linear warmup\\
Warmup steps & First 6\% of steps & First 6\% of steps & First 10\% of steps\\
Weight decay & 0.01 & 0.01 & 0.01\\
Attention dropout & 0.1 & 0.1 & 0.1\\
Dropout & 0.1 & 0.1 & 0.1\\
Early stopping criterion & No improvements  & No improvements & None\\
& over 5\% of steps & over 2.5\% of steps & \\
\bottomrule
\end{tabular}%
\caption{Hyperparameters for fine-tuning.}
\label{table:hyperparams_finetuning}
\end{center}
\end{table*}

\section*{Appendices}
\label{sec:appendix}

\section{Masked $n$ Characters Prediction}
The task of predicting first or last $n$ characters of a masked token ($n$ Chars) is trained with the token-level cross-entropy loss averaged over the masked ones only, following First Char \cite{yamaguchi-etal-2021-frustratingly}. We mask 15\% of input tokens the same as in BERT.

Our method generates a label dictionary for $n$ Chars similar to that of First Char. We used a pretrained tokenizer of RoBERTa (\texttt{roberta-base}) provided by the Transformers library and generated a label of each token in the vocabulary by picking out a specified number of characters. We did not count a special blank character of Ġ when generating labels. The average number of characters per token in the \texttt{roberta-base} vocabulary is $5.72 \pm 1.78$.

\section{Detailed Experimental Setup} \label{sec:hyperparameters}
We trained our models with four NVIDIA Tesla V100 (32GB) for pretraining and one for fine-tuning.
Note that we used eight V100 GPUs for MLM to match the total batch size of 128 used for other models.

\subsection{Pretraining}
We pretrained all models for 500k steps following \citet{alajrami-aletras-2022-pre} and optimized the models with AdamW \cite{loshchilov2018decoupled}. Table \ref{table:hyperparams_pretraining} shows the hyperparameter settings used in pretraining.

\subsection{Fine-tuning}
Table \ref{table:hyperparams_finetuning} lists the hyperparameters for fine-tuning models on GLUE, SQuAD, and UD benchmarks. For GLUE and SQuAD, we used early stopping.
For UD, we compute an average for each token over the top four layers of the BERT hidden representations and use it as an input to BAP. The dimensionalities of arc and relation features given to each biaffine module are 500 and 100, respectively.

\section{GLUE}
\subsection{Evaluation Metrics}
Following previous work \cite{aroca-ouellette-rudzicz-2020-losses}, we report matched accuracy for MNLI \cite{williams-etal-2018-broad}, Matthews correlation for CoLA \cite{warstadt-etal-2019-neural}, Spearman correlation for STS-B \cite{cer-etal-2017-semeval}, accuracy for MRPC \cite{dolan-brockett-2005-automatically}, F1 scores for QQP\footnote{\url{https://quoradata.quora.com/First-Quora-Dataset-Release-Question-Pairs}} and SQuAD, and accuracy for all other tasks, including SST-2 \cite{socher-etal-2013-recursive}, QNLI \cite{wang2018glue} and RTE \cite{10.1007/11736790_9,barhaim2006second,giampiccolo-etal-2007-third,bentivogli2009fifth}.

\subsection{Results} \label{sec:glue}
Table \ref{table:base_result_glue_std} shows the detailed GLUE results with standard deviations for each mean value. We can see that standard deviations on low-resource corpora with fewer than 10k samples tend to be larger than that of high-resource corpora.

\renewcommand*{\arraystretch}{1.0}
\begin{table*}[!t]
\begin{center}
\small
\resizebox{\linewidth}{!}{%
\begin{tabular}{lcccccccc}
\toprule
 & \textbf{MNLI} & \textbf{QQP} & \textbf{QNLI} & \textbf{SST} & \textbf{CoLA} & \textbf{STS} & \textbf{MRPC} & \textbf{RTE}\\ 
 \textbf{Model} & 393k & 364k & 105k & 67k & 8.6k & 5.7k & 3.7k & 2.5k\\ 
\midrule
MLM & 82.3 (0.3) & 86.9 (0.2) & 89.2 (0.1) & 91.8 (0.4) & 58.0 (2.0) & 87.0 (0.4) & 86.7 (0.4) & 64.8 (1.0)\\
\midrule
First 9 Chars & 81.6 (0.2) & 86.4 (0.3) & 89.2 (0.4) & 91.9 (0.7) & 53.0 (2.5) & 85.6 (0.5) & 85.2 (1.1) & 58.2 (4.5)\\
First 5 Chars & 82.0 (0.2) & 86.6 (0.1) & 89.3 (0.3) & 91.1 (0.3) & 51.8 (2.4) & 85.6 (0.3) & 85.5 (0.2) & 59.2 (2.2)\\
First 4 Chars & 82.0 (0.2) & 86.6 (0.2) & 89.6 (0.3) & 91.3 (0.4) & 54.2 (2.0) & 85.5 (0.4) & 85.7 (0.7) & 57.3 (3.1)\\
First 3 Chars & 81.9 (0.3) & 86.8 (0.2) & 88.7 (0.7) & 90.7 (0.5) & 52.0 (1.2) & 85.9 (0.4) & 85.6 (0.5) & 58.9 (2.2)\\
First 2 Chars & 81.1 (0.4) & 86.5 (0.1) & 88.6 (0.4) & 90.8 (0.4) & 51.1 (1.0) & 85.1 (0.3) & 83.7 (1.0) & 60.6 (4.8)\\
First 1 Char & 80.5 (0.5) & 86.3 (0.1) & 88.5 (0.2) & 90.4 (0.2) & 48.6 (2.0) & 84.7 (1.0) & 83.3 (2.0) & 60.0 (1.4)\\
First Char & 80.7 (0.3) & 86.3 (0.2) & 88.2 (0.3) & 90.6 (0.3) & 50.0 (1.7) & 85.1 (0.2) & 85.4 (0.7) & 59.5 (1.1)\\
\midrule
Last 9 Chars & 82.1 (0.1) & 86.7 (0.3) & 89.3 (0.3) & 91.4 (0.7) & 55.0 (1.3) & 85.6 (0.3) & 85.1 (0.9) & 57.6 (2.2)\\
Last 5 Chars & 81.8 (0.1) & 86.4 (0.1) & 89.1 (0.3) & 91.3 (0.6) & 54.8 (1.8) & 85.8 (0.3) & 85.4 (1.0) & 58.7 (4.8)\\
Last 4 Chars & 81.6 (0.2) & 86.6 (0.1) & 89.4 (0.2) & 90.2 (0.5) & 56.0 (0.7) & 85.6 (0.7) & 86.2 (0.5) & 56.9 (2.5)\\
Last 3 Chars & 81.3 (0.2) & 86.4 (0.4) & 88.9 (0.3) & 91.0 (0.7) & 53.0 (2.0) & 84.9 (0.2) & 84.9 (1.0) & 56.1 (3.7)\\
Last 2 Chars & 81.0 (0.4) & 86.3 (0.3) & 88.0 (0.3) & 90.7 (0.5) & 50.7 (2.6) & 84.5 (0.6) & 85.7 (0.9) & 58.6 (2.1)\\
Last 1 Char & 80.2 (0.2) & 86.3 (0.2) & 87.9 (0.4) & 90.4 (0.4) & 54.8 (1.8) & 84.6 (0.4) & 84.8 (1.6) & 61.2 (0.8)\\
Last Char & 79.8 (0.3) & 86.0 (0.1) & 87.5 (0.1) & 90.2 (0.3) & 48.8 (2.6) & 85.2 (0.3) & 85.2 (0.6) & 55.7 (3.4)\\
\bottomrule
\end{tabular}%
}
\caption{Results on GLUE dev sets with standard deviations over five runs in parentheses. Values under dataset names are the number of their corresponding training samples.}
\label{table:base_result_glue_std}
\end{center}
\end{table*}

\section{Relative Performance Difference from MLM} \label{sec:performance_degradation}
Table \ref{table:base_result_performance_degradation} shows the results on GLUE, SQuAD, and UD benchmarks along with the relative performance differences from MLM in percent.

\section{License}
SQuAD and UD (EN-EWT) are distributed under the CC BY-SA 4.0 license. GLUE has different licenses but is freely available for typical machine learning uses. We used all of the corpora for benchmarking purposes only and did not modify their contents.

\renewcommand*{\arraystretch}{1.0}
\begin{table*}[t!]
\begin{center}
\tabcolsep 4pt
\resizebox{\linewidth}{!}{%
\begin{tabular}{lcccccccc|c|c|c}
\toprule
 & \textbf{MNLI} & \textbf{QQP} & \textbf{QNLI} & \textbf{SST} & \textbf{CoLA} & \textbf{STS} & \textbf{MRPC} & \textbf{RTE} & \textbf{GLUE Avg.} & \multicolumn{1}{c|}{\textbf{SQuAD}} & \multicolumn{1}{c}{\textbf{UD}}\\ 
 \textbf{Model} & 393k & 364k & 105k & 67k & 8.6k & 5.7k & 3.7k & 2.5k &  & \multicolumn{1}{c|}{88k} & \multicolumn{1}{c}{13k} \\ 
\midrule
MLM & 82.3 & 86.9 & 89.2 & 91.8 & 58.0 & 87.0 & 86.7 & 64.8 & 80.8 & 88.1 & 88.8\\
\midrule
\textbf{First}\\
9 Chars & 81.6 (-0.9) & 86.4 (-0.6) & 89.2 (0.0) & 91.9 (+0.1) & 53.0 (-8.7) & 85.6 (-1.7) & 85.2 (-1.6) & 58.2 (-10.2) & 78.9 (-2.4) & 87.4 (-0.9) & 88.5 (-0.3)\\
5 Chars & 82.0 (-0.4) & 86.6 (-0.3) & 89.3 (+0.1) & 91.1 (-0.7) & 51.8 (-10.8) & 85.6 (-1.6) & 85.5 (-1.4) & 59.2 (-8.7) & 78.9 (-2.4) & 87.9 (-0.3) & 88.5 (-0.4)\\
4 Chars & 82.0 (-0.4) & 86.6 (-0.4) & 89.6 (+0.4) & 91.3 (-0.5) & 54.2 (-6.7) & 85.5 (-1.7) & 85.7 (-1.1) & 57.3 (-11.7) & 79.0 (-2.3) & 87.9 (-0.2) & 88.8 (0.0)\\
3 Chars & 81.9 (-0.5) & 86.8 (-0.2) & 88.7 (-0.5) & 90.7 (-1.2) & 52.0 (-10.5) & 85.9 (-1.2) & 85.6 (-1.2) & 58.9 (-9.1) & 78.8 (-2.5) & 87.6 (-0.6) & 88.1 (-0.8)\\
2 Chars & 81.1 (-1.5) & 86.5 (-0.5) & 88.6 (-0.6) & 90.8 (-1.1) & 51.1 (-12.0) & 85.1 (-2.2) & 83.7 (-3.4) & 60.6 (-6.5) & 78.4 (-3.0) & 86.8 (-1.5) & 87.8 (-1.1)\\
1 Char & 80.5 (-2.2) & 86.3 (-0.7) & 88.5 (-0.8) & 90.4 (-1.5) & 48.6 (-16.3) & 84.7 (-2.7) & 83.3 (-3.8) & 60.0 (-7.5) & 77.8 (-3.8) & 86.1 (-2.3) & 87.6 (-1.4)\\
First Char & 80.7 (-2.0) & 86.3 (-0.7) & 88.2 (-1.1) & 90.6 (-1.3) & 50.0 (-13.8) & 85.1 (-2.2) & 85.4 (-1.5) & 59.5 (-8.2) & 78.2 (-3.2) & 85.6 (-2.9) & 87.8 (-1.2)\\
\midrule
\textbf{Last}\\
9 Chars & 82.1 (-0.3) & 86.7 (-0.2) & 89.3 (+0.1) & 91.4 (-0.4) & 55.0 (-5.2) & 85.6 (-1.7) & 85.1 (-1.8) & 57.6 (-11.1) & 79.1 (-2.2) & 88.4 (+0.3) & 88.7 (-0.2)\\
5 Chars & 81.8 (-0.6) & 86.4 (-0.6) & 89.1 (-0.1) & 91.3 (-0.6) & 54.8 (-5.6) & 85.8 (-1.4) & 85.4 (-1.4) & 58.7 (-9.5) & 79.2 (-2.1) & 87.5 (-0.7) & 88.5 (-0.4)\\
4 Chars & 81.6 (-0.9) & 86.6 (-0.3) & 89.4 (0.3) & 90.2 (-1.7) & 56.0 (-3.5) & 85.6 (-1.7) & 86.2 (-0.5) & 56.9 (-12.2) & 79.1 (-2.2) & 87.6 (-0.6) & 88.4 (-0.4)\\
3 Chars & 81.3 (-1.2) & 86.4 (-0.6) & 88.9 (-0.4) & 91.0 (-0.9) & 53.0 (-8.7) & 84.9 (-2.4) & 84.9 (-2.0) & 56.1 (-13.5) & 78.3 (-3.1) & 87.0 (-1.3) & 88.4 (-0.5)\\
2 Chars & 81.0 (-1.6) & 86.3 (-0.7) & 88.0 (-1.3) & 90.7 (-1.2) & 50.7 (-12.7) & 84.5 (-3.0) & 85.7 (-1.1) & 58.6 (-9.6) & 78.2 (-3.3) & 87.0 (-1.3) & 88.0 (-1.0)\\
1 Char & 80.2 (-2.5) & 86.3 (-0.8) & 87.9 (-1.4) & 90.4 (-1.5) & 54.8 (-5.5) & 84.6 (-2.7) & 84.8 (-2.1) & 61.2 (-5.6) & 78.8 (-2.5) & 86.0 (-2.4) & 87.8 (-1.1)\\
Last Char & 79.8 (-3.0) & 86.0 (-1.1) & 87.5 (-1.9) & 90.2 (-1.8) & 48.8 (-15.9) & 85.2 (-2.1) & 85.2 (-1.7) & 55.7 (-14.1) & 77.3 (-4.4) & 85.5 (-3.0) & 88.1 (-0.8)\\
\bottomrule
\end{tabular}%
}
\caption{Results on GLUE, SQuAD, and UD datasets along with their relative performance differences from MLM in percent.}
\label{table:base_result_performance_degradation}
\end{center}
\end{table*}

\end{document}